\documentclass[12pt]{article}
\usepackage{amsmath}
\usepackage{times}
\usepackage{graphicx}
\usepackage{color}
\usepackage{amsfonts}
\usepackage{multirow}
\usepackage{subfig}
\usepackage[authoryear]{natbib}
\usepackage{amsthm}
\theoremstyle{postulate}
\newtheorem{pot}{Postulate}	

\theoremstyle{definition}
\newtheorem{dfn}{Definition}
\usepackage{slashbox}
\usepackage{rotating}
\usepackage{bbm}
\usepackage{latexsym}

\newcommand{\captionfonts}{\normalsize}

\makeatletter  
\long\def\@makecaption#1#2{%
  \vskip\abovecaptionskip
  \sbox\@tempboxa{{\captionfonts #1: #2}}%
  \ifdim \wd\@tempboxa >\hsize
    {\captionfonts #1: #2\par}
  \else
    \hbox to\hsize{\hfil\box\@tempboxa\hfil}%
  \fi
  \vskip\belowcaptionskip}
\makeatother   
\newcommand\independent{\protect\mathpalette{\protect\independenT}{\perp}}
    \def\independenT#1#2{\mathrel{\setbox0\hbox{$#1#2$}%
   \copy0\kern-\wd0\mkern4mu\box0}}

\begin{document}
\hspace{13.9cm}1

\ \vspace{20mm}\\
{\centering
{\LARGE  Causal Inference on Discrete Data via Estimating Distance Correlations}

\ \\
{\bf \large Furui Liu, Laiwan Chan}\\
{Department of Computer Science and Engineering,}\\
{The Chinese University of Hong Kong}\\
%

}
\thispagestyle{empty}
\markboth{}{NC instructions}
\ \vspace{-0mm}\\
%
\begin{center} {\bf Abstract} \end{center}
In this paper, we deal with the problem of inferring  causal directions when the data is on discrete domain. By considering the distribution of the cause $P(X)$ and the conditional distribution mapping cause to effect $P(Y|X)$ as independent random variables, we propose to infer the causal direction via comparing the distance correlation between $P(X)$ and $P(Y|X)$ with the distance correlation between $P(Y)$ and $P(X|Y)$. We infer ``$X$ causes $Y$'' if the dependence coefficient between $P(X)$ and $P(Y|X)$ is smaller. Experiments are performed to show the performance of the proposed method.  \\\\
{\bf Editor}: Aapo Hyv\"{a}rinen
 

\section{Introduction}
Inferring the causal direction between two variables from observational data becomes a hot research topic. Additive noise models (ANMs) \citep{additive1,lingam,hoyer2009nonlinear,peters2011causal,hoyer2009,zhang2009identifiability,directlingam,pwlingam,estimationSEM,hyttinen2012learning,additive3} are preliminary trials to solve this problem. They assume that the effect is governed by the cause and an additive noise, and the causal inference is done by finding the  direction that admits such a model. Recently, under another view of exploiting the asymmetry between cause and effect, the linear trace method (LTr) \citep{zscheischler2011testing,janzing2010telling} and information geometric causal inference (IGCI) \citep{janzing2012information} are proposed. Suppose $X$ is the cause and $Y$ is the effect. Based on the fact that the generating of $P(X)$ is independent with that of $P(Y|X)$ \citep{janzing2010causal,lemeire2013replacing,scholkopf2012causal}, LTr suggests that the trace condition is fulfilled in the causal direction while violated in the anti-causal direction, and IGCI shows that the density of the cause and the log slope of the function transforming cause to effect are uncorrelated while the density of the effect and the log slope of the inverse of the function are positively correlated. By accessing these so-called cause-effect asymmetries, one can determine the causal direction. Then a kernel method using the framework of IGCI to deal with high dimensional variables is developed \citep{chen2014causal}, and nonlinear extensions of trace method is presented \citep{chen2013nonlinear}.   

In some situations, the variables of interest are on discrete domains, and researchers have adopted additive noise models to discrete data for causal inference \citep{peters2010,peters2011causal}. Given observations of the variable pair, they do regressions in both directions, and test the independence between the residuals and the regressors. The direction that admits an additive noise model is inferred as the causal direction. However,  ANMs may not be suitable for modeling discrete variables in some situations. For example, it is not natural to adopt ANMs to modeling categorical variables. Methods with wider applicability would be valuable for causal inference on discrete data. 

Motivated by the postulate that the generating of $P(X)$ is independent with that of $P(Y|X)$, we suggest that the $(P(x),P(Y|x))$ is an observation of a pair of variables that are independent with each other (here $P(x)=P(X=x)$, referring to the probability at one specified point). To infer the causal direction, we calculate the dependence coefficients between $P(X)$ and $P(Y|X)$, and $P(Y)$ and $P(X|Y)$. Then the direction that induces smaller correlation is inferred as the causal direction. Without a functional causal model assumption, our method is with wider applicability than traditional ANMs.  Various experiments are conducted to demonstrate the performance of the proposed method.

This paper is organized as follows. Section 2 defines the problem. Section 3 presents the causal inference principle. Section 4 gives the detailed causal inference method. Section 5 shows the experiments, and section 6 concludes the whole paper.

\section{Problem Description}
Suppose we have a set of observed variables $X$ and $Y$ with support domain $\mathcal{X}$ and $\mathcal{Y}$ respectively. $X$ is the cause and $Y$ is the effect but we do not have prior knowledge about the causal direction.  We assume that they are on discrete domain (for continuous variable we can perform discretization) and there are no latent confounders. We want to identify the causal direction (``X causes Y'' or ``Y causes X''). For clarity, we list the symbols that may appear in the following sections in table 1. Since we constrain variables in finite range,  $P(X,Y)$ and $P(Y|X)$ can be written as  matrices, and  $P(Y|x)$ is a vector in the matrix $P(Y|X)$. We  use these representations in the rest of the paper.
 \begin{table}[h]
       \centering
         \caption{Lookup table}
        \begin{tabular}{c|c}
        \hline
        Symbol & Description \\\hline
        
        $\mathcal{X}$ & support of variable $X$ \\
        $\mathcal{Y}$ & support of variable $Y$ \\
        $P(X)$ & distribution of $X$\\
        $P(x)$ & probability of $X=x$\\ 
                $P(X,Y)$ & joint distribution of  $(X,Y)$ \\
           $P(Y|X)$ & conditional distribution of $Y$ given $X$ \\
           $P(Y|x)$ & conditional distribution of $Y$ given $X=x$\\
            $|\cdot|$ & cardinality of a set\\\hline
        \end{tabular}     
        \end{table}

\section{Causal Inference Principle}
In this section, we will present the principle that we are using for causal inference on discrete data. We start with the basic idea.
\subsection{Basic Idea}
The basic idea  is to consider the $(P(x), P(Y|x))$ as a realization of a variable pair, and the two variables (one is one dimensional and the other is high dimensional) are independent with each other. See figure 1 for an example.  
\begin{figure}[h]
\centering
\includegraphics[width=0.6\textwidth]{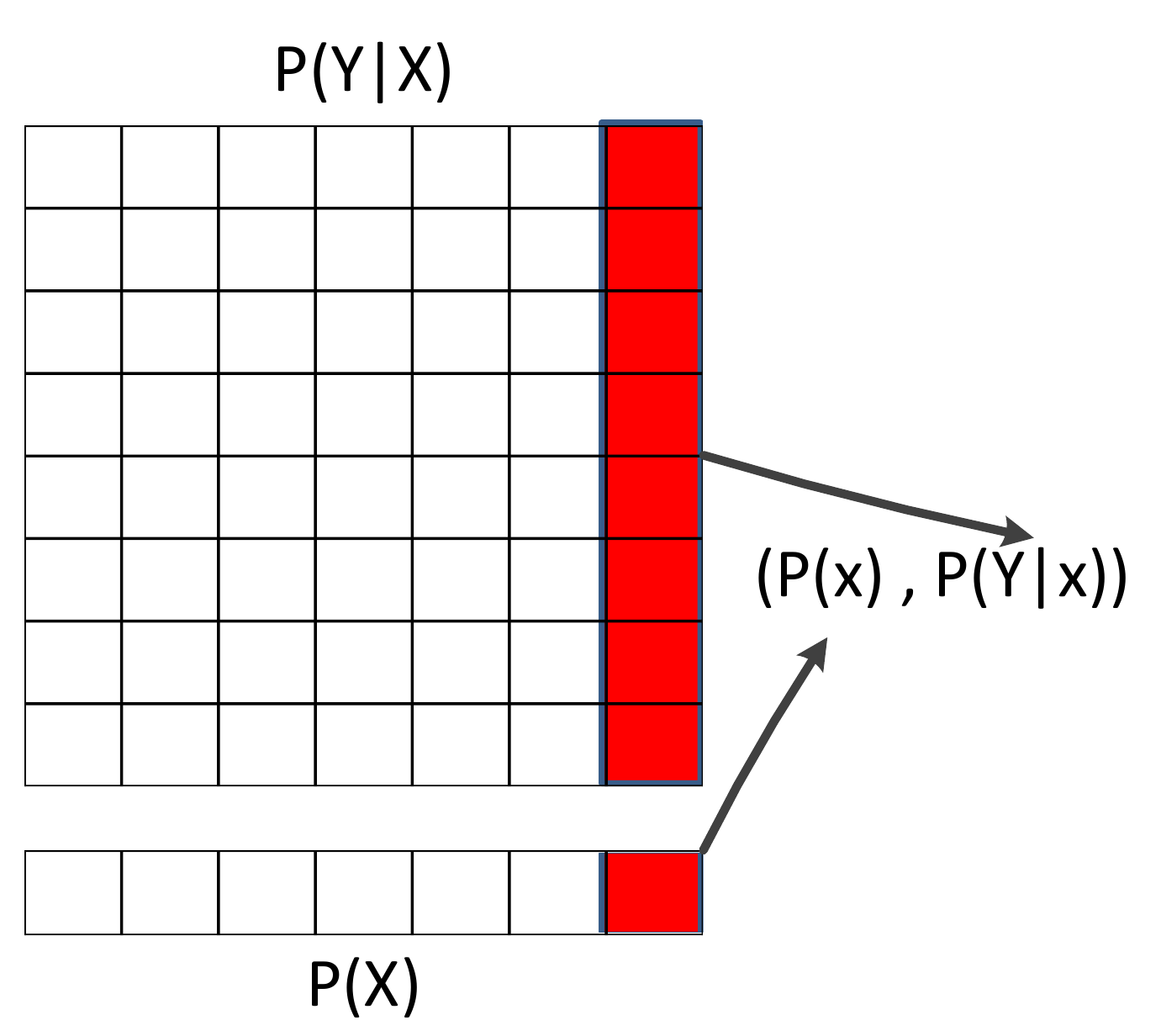}
\caption{$P(X)$ and $P(Y|X)$ as random variables}
\end{figure}

Figure 1 shows an example of $P(X)$ and $P(Y|X)$. Suppose $|\mathcal{X}| = M$ and $|\mathcal{Y}| = L$ (here $M=7$ and $L=8$). $P(X)$ is a vector in $R^M$ and  $P(Y|X)$ is a matrix in $R^{L\times M}$. The highlighted (red bars) grids are a pair $(P(x), P(Y|x))$. Consider $P(X)$ and $P(Y|X)$ as two independent random variables. The generating of the $P(X)$ and $P(Y|X)$ is done by drawing realizations $(P(x), P(Y|x))$ at each possible value of $x$ (shifting the red bars from right to left). We have $|\mathcal{X}|$  realizations. We formalize this in postulate 1.
\begin{pot}
$P(X)$ and $P(Y|X)$ are both random variables taking realizations at different $x$. $P(X)$ is independent with $P(Y|X)$.
\end{pot}
Once we have this postulate in mind, one could seek for some properties induced by it for causal discovery. We will discuss this in the next section. 
\subsection{Distance Correlation}
If we want to characterize the dependence between $P(X)$ and $P(Y|X)$, one measurement  is the correlation. However, $P(Y|X)$ is a high dimensional random vector. Adopting traditional dependence coefficients like Pearson correlations would cause certain estimation bias when sample size is not large. Moreover, it would be useful if the independence between variables corresponds to 0 correlation. This is not true if we use traditional correlations. Here we propose to use distance correlation \citep{szekely2007measuring} as the dependence measurement. 

Distance correlation is a measurement of dependence between two random variables (one dimensional or high dimensional). Suppose we have two random variables $(\alpha,\beta)$, with characteristic functions $f_\alpha(t)$ and $f_\beta(s)$ respectively. Their joint characteristic function  is $f_{\alpha,\beta}(t,s)$. The distance covariance is defined as below. 

\begin{dfn}
The distance covariance $\mathcal{C}(\alpha,\beta)$ between two random variables $(\alpha,\beta)$  is 
\begin{equation}
\mathcal{C}^2(\alpha,\beta) = \|f_{\alpha,\beta}(t,s) - f_\alpha(t)f_\beta(s)\|^2
\end{equation}
\end{dfn}
Here $\|\cdot\|$ refers the weighted $L_2$ norm, and similarly we can define distance variance $\mathcal{C}^2(\alpha,\alpha)$ \citep{szekely2007measuring}. Then the distance correlation is defined as
\begin{dfn}
The distance correlation $\mathcal{D}(\alpha,\beta)$ is 
\begin{equation}
\mathcal{D}(\alpha,\beta) = \frac{\mathcal{C}(\alpha,\beta) }{\sqrt{\mathcal{C}(\alpha,\alpha)\mathcal{C}(\beta,\beta)}}
\end{equation}
and $\mathcal{D}(\alpha,\beta) = 0$ if $\mathcal{C}(\alpha,\alpha)=0$ or  $\mathcal{C}(\beta,\beta)=0$.
\end{dfn}
This dependence measurement is a distance metric between the joint and multiply of the marginal characteristic functions. There are other methods to measure the dependence, like mutual information and kernel independence measurements \citep{gretton2005kernel}. However, mutual information is hard to estimate given finite sample size. Kernel methods involve a few parameters (kernel functions, kernel widths) which is not easy to choose. So we use this metric in our paper. Then we discuss how to estimate the distance correlation empirically from data \citep{szekely2007measuring}. Suppose we have $n$ observations of two random variables $(\alpha,\beta)$ as $\{(\alpha_i,\beta_i)\}_{i=1}^n$. For variable $\alpha$ and $\beta$, we can construct
\begin{equation}
a_{ij}  =  \|\alpha_i - \alpha_j\|,~~~~
 a_{i\cdot}  =  \frac{1}{n}\sum_{j=1}^{n}a_{ij},~~~~
  a_{\cdot j}  =  \frac{1}{n}\sum_{i=1}^{n}a_{ij},~~~~
   a_{\cdot \cdot}  =  \frac{1}{n^2}\sum_{i,j=1}^{n}a_{ij}
\end{equation}
\begin{equation}
b_{ij} = \|\beta_i - \beta_j\|,~~~~
 b_{i\cdot} = \frac{1}{n}\sum_{j=1}^{n}b_{ij},~~~~
  b_{\cdot j} = \frac{1}{n}\sum_{i=1}^{n}b_{ij},~~~~
   b_{\cdot \cdot} = \frac{1}{n^2}\sum_{i,j=1}^{n}b_{ij}
\end{equation}
and then we construct matrices $A$ and $B$, with its entries as
\begin{equation}
A_{ij} = a_{ij} - a_{i\cdot} -  a_{\cdot j} + a_{\cdot \cdot}
\end{equation}
\begin{equation}
B_{ij} = b_{ij} - b_{i\cdot} -  b_{\cdot j} + b_{\cdot \cdot}
\end{equation}
 Then  we can estimate the empirical distance covariance as follows \citep{szekely2007measuring}.
\begin{dfn}
The empirical distance covariance $\mathcal{C}_n(\alpha,\beta)$ is 
\begin{equation}
\mathcal{C}_n(\alpha,\beta) = \frac{1}{n}\sqrt{\sum_{i,j=1}^{n}A_{ij}B_{ij}}
\end{equation}
\end{dfn}

We can estimate the empirical distance correlation using the empirical distance covariance. The distance correlation has a property that $\mathcal{D}(\alpha,\beta) = 0$ implies  independence between $\alpha$ and $\beta$. We show that this helps to identify the causal direction in the next section.

\subsection{Inferring Causal Directions}
In this section we discuss how to infer the causal directions. Suppose we have the joint distribution of the variable pair as $P(X,Y)$. We are able to factorize it in two directions and get $(P(X),P(Y|X))$ and $(P(Y),P(X|Y))$. Each of them is a random variable pair. We define the dependence measurements of them as below.
\begin{dfn}
The dependence measurements are defined as
\begin{eqnarray}
\mathcal{D}_{X\to Y} & = & \mathcal{D}(P(X),P(Y|X)) \\
\mathcal{D}_{Y\to X} & = & \mathcal{D}(P(Y),P(X|Y))
\end{eqnarray}
\end{dfn}

If postulate 1 is accepted,  then in the causal direction, the distance correlation between $P(X)$ and $P(Y|X)$ reaches the lower bound as
\begin{equation}
\mathcal{D}_{X\to Y} = 0
\end{equation}
Since the distance correlation is nonnegative, in the anti-causal direction we have
\begin{equation}
\mathcal{D}_{Y\to X} \geq 0
\end{equation}
and now we get the causal inference principle as
\begin{equation}
\mathcal{D}_{Y\to X} \geq \mathcal{D}_{X\to Y}
\end{equation}
Intuitively speaking, in the causal direction we get smaller dependence coefficient between the marginal distribution and the conditional distribution than that in the anti-causal direction. One thing worth attention is that the domain size should be reasonably large to generate reliable statistics. We will give the detailed causal inference method in the next section.
\section{Causal Inference Method}
In this section we give a causal inference method which identifies the causal direction via estimating the distance correlations. If X causes Y, the $\mathcal{D}_{X\to Y}$ should be smaller than $\mathcal{D}_{Y\to X}$. However, estimating the coefficients from samples may induce random errors. So we introduce a threshold $\epsilon$. They are significantly different if their difference is larger than  $\epsilon$, and we can decide the causal direction. Otherwise we stay undecided. We detail the inference method in table 2. 
\begin{table}[h]
\caption{Causal inference method}
\centering
\begin{tabular}{p{\textwidth}}\hline\hline
  \textbf{Algorithm 1: Causal inference via estimating distance correlations (DC)}  \\ \hline
   Input: sample of the discrete variables $X$ and $Y$, threshold $\epsilon$  \\
 1.  Construct the vector recording the distribution $P(X)$ and the matrix recording the conditional distribution $P(Y|X)$.  Calculate $\mathcal{D}_{X\to Y}  =  \mathcal{D}(P(X),P(Y|X))$.       \\
 2.  Construct the vector recording the distribution $P(Y)$ and the matrix recording the conditional distribution $P(X|Y)$.  Calculate $\mathcal{D}_{Y\to X}  =  \mathcal{D}(P(Y),P(X|Y))$.  \\
 3.  Decide the causal direction: \\
 If $\mathcal{D}_{Y\to X} - \mathcal{D}_{X\to Y} > \epsilon$, output ``$X$ causes $Y$''.\\
        If $\mathcal{D}_{X\to Y} - \mathcal{D}_{Y\to X} > \epsilon$, output ``$Y$ causes $X$''.\\
         Else, output ``No decision made''.       \\\hline \hline
\end{tabular}
\end{table}

     From table 2, one could see that our  method identifies the cause and the effect by factorizing the joint distribution in two directions and comparing the dependence coefficients ($\mathcal{D}_{X\to Y}$ and $\mathcal{D}_{Y\to X}$). The one with smaller distance correlation (between the marginal distribution and the conditional distribution) is inferred as the causal direction. We name it causal inference via estimating distance correlations (abbreviated as DC). Next we analyze the computational cost of our method. Suppose the sample size is $n$. The time for constructing the matrix recording the joint distribution is $O(n)$, and the times for calculating $\mathcal{D}_{X\to Y}$ and $\mathcal{D}_{Y\to X}$ are $O(|\mathcal{X}|^2)$ and  $ O(|\mathcal{Y}|^2)$ respectively. So the total time is $O(n+|\mathcal{X}|^2 + |\mathcal{Y}|^2)$. One can see that this method is of low computational complexity (linear with respect to sample size). This would be verified in experiments.

           \section{Experiments}
             In this section we test the performance of our method (DC).  The compared algorithm is the discrete regression (DR) \citep{peters2011causal}. We perform experiments under various settings. Section 5.1 shows the performance of DC on identifying ANMs with different $\mathcal{N}$. Section 5.2 presents the performance of DC when the distribution of the cause and the conditional distribution mapping cause to effect are randomly generated. Section 5.3 tests the efficiency of the algorithms. Section 5.4 discusses the choice  of the threshold parameter $\epsilon$. Section 5.5 shows the performance of DC at different decision rates. In section 5.6, we apply DC to real world cause-effect pairs (with discretization) to show its capability in solving practical problems. 
             \subsection{Additive Noise Models}
             We first evaluate the accuracies of DC on identifying ANMs with different ranges of noise term $N$. The model is written as
             \begin{equation}
             Y=f(X)+N,N\independent X
             \end{equation}
             The function $f$ is constructed by random mapping from $\mathcal{X}=\{1,2,...,30\}$ to  $\mathcal{Y}_0=\{1,2,...,30\}$.  Suppose the support of the noise is $\mathcal{N}$. The noise domain $\mathcal{N}$ is chosen as:
             \begin{enumerate}
              \item $\mathcal{N} = \{0,1\}$.
              \item $\mathcal{N} = \{-1,0,1\}$.
             \item $\mathcal{N} = \{-2,-1,0,1,2\}$.
             \item $\mathcal{N} = \{-3,-2,-1,0,1,2,3\}$.
             \end{enumerate}
             The probability distributions of the cause $X$ are chosen by: (1) randomly generate a  vector (length $|\mathcal{X}|$) with each entry being an integer between $[1,|\mathcal{X}|/4]$. (2) Normalize it to unit sum.  We generate the probability distributions of the noise $N$ using the same way.  In each trial, the algorithms are forced to make a decision.  For each noise setting, we randomly generate 500 functions. Thus we have 500 additive noise models. $\mathcal{Y}$  for each model could be different due to the randomness of the mappings.  Then we sample  200, 300, 500, 1000, 2000, 4000 points for each model, and apply DC and DR to the samples. The plots showing the accuracies of the algorithms are given in figure 2.
             \begin{figure}[h]%
                 \centering
                  \subfloat[ $|\mathcal{N}|=2$]{{\includegraphics[width=.5\textwidth]{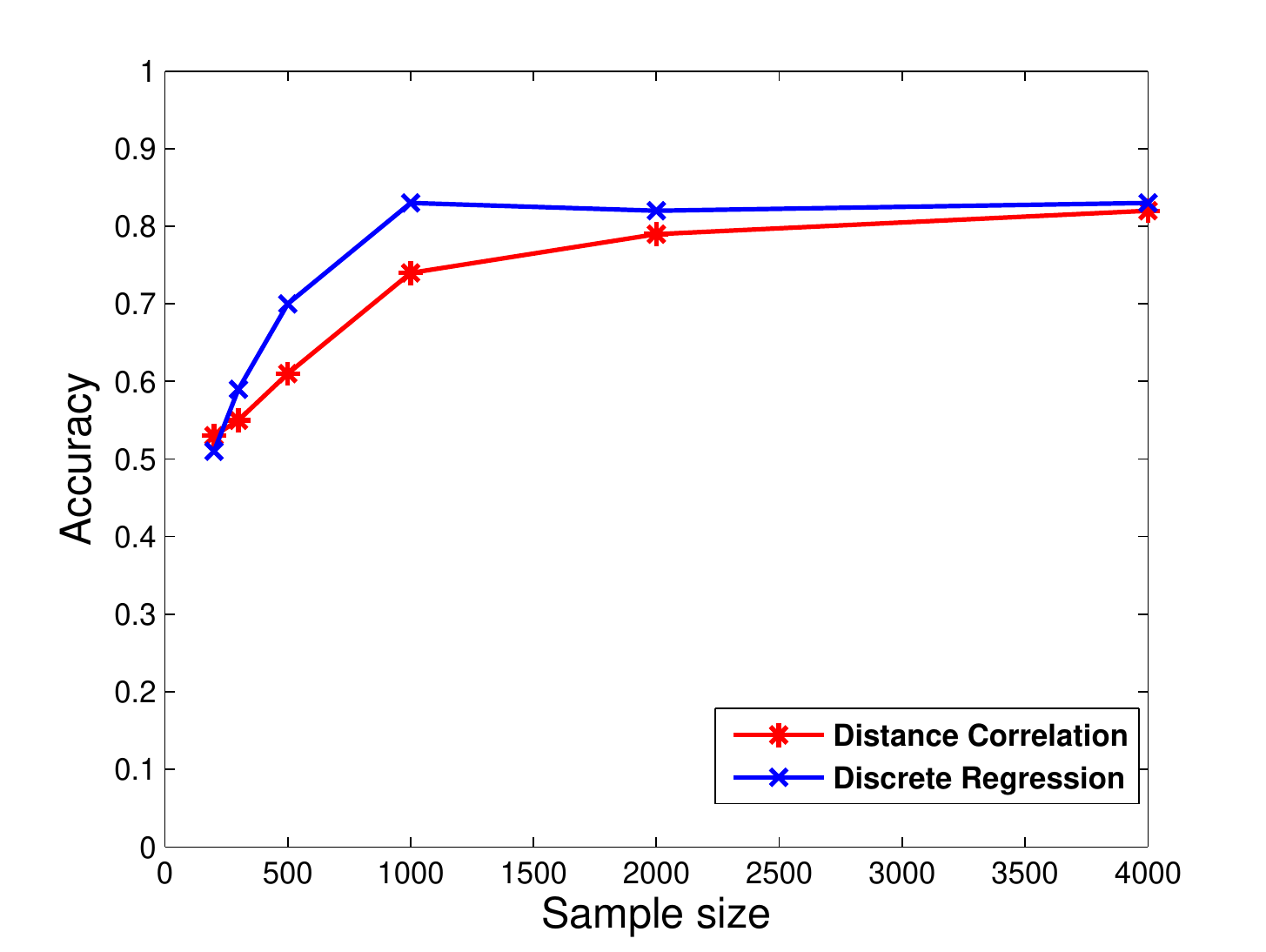} }}
                    \subfloat[ $|\mathcal{N}|=3$]{{\includegraphics[width=.5\textwidth]{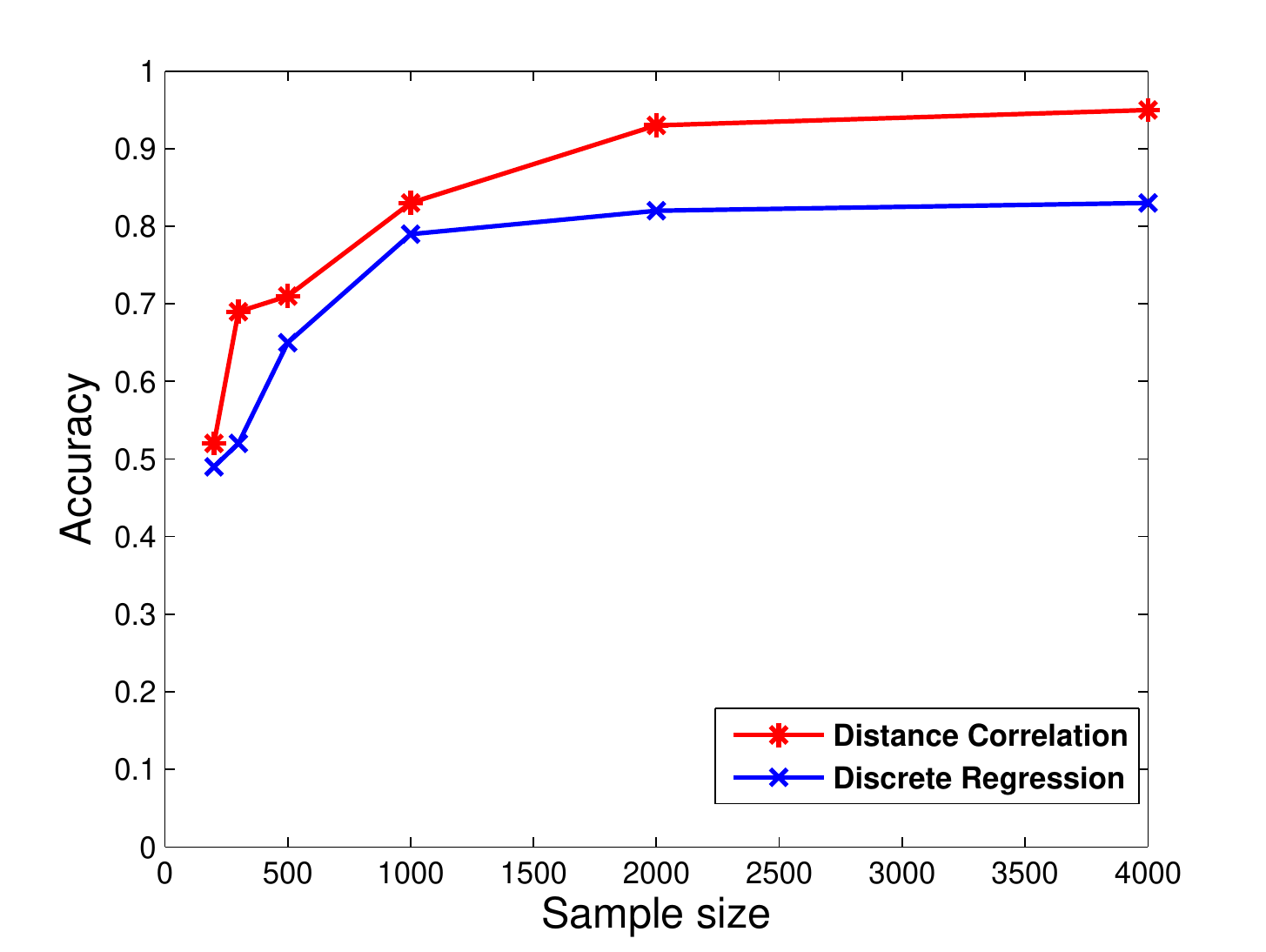} }}  
                    
                       \subfloat[ $|\mathcal{N}|=5$]{{\includegraphics[width=.5\textwidth]{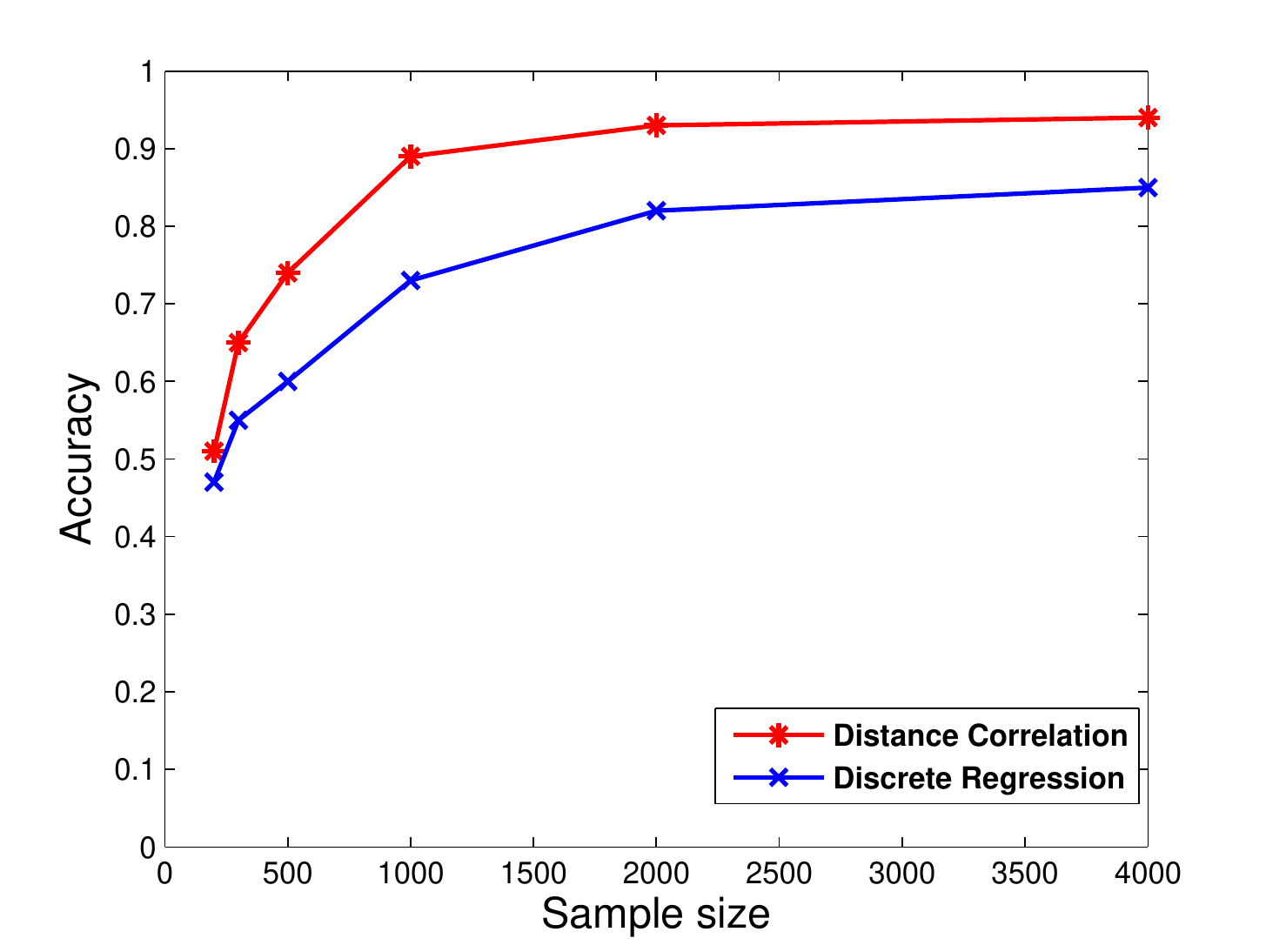} }}
                                          \subfloat[ $|\mathcal{N}|=7$]{{\includegraphics[width=.5\textwidth]{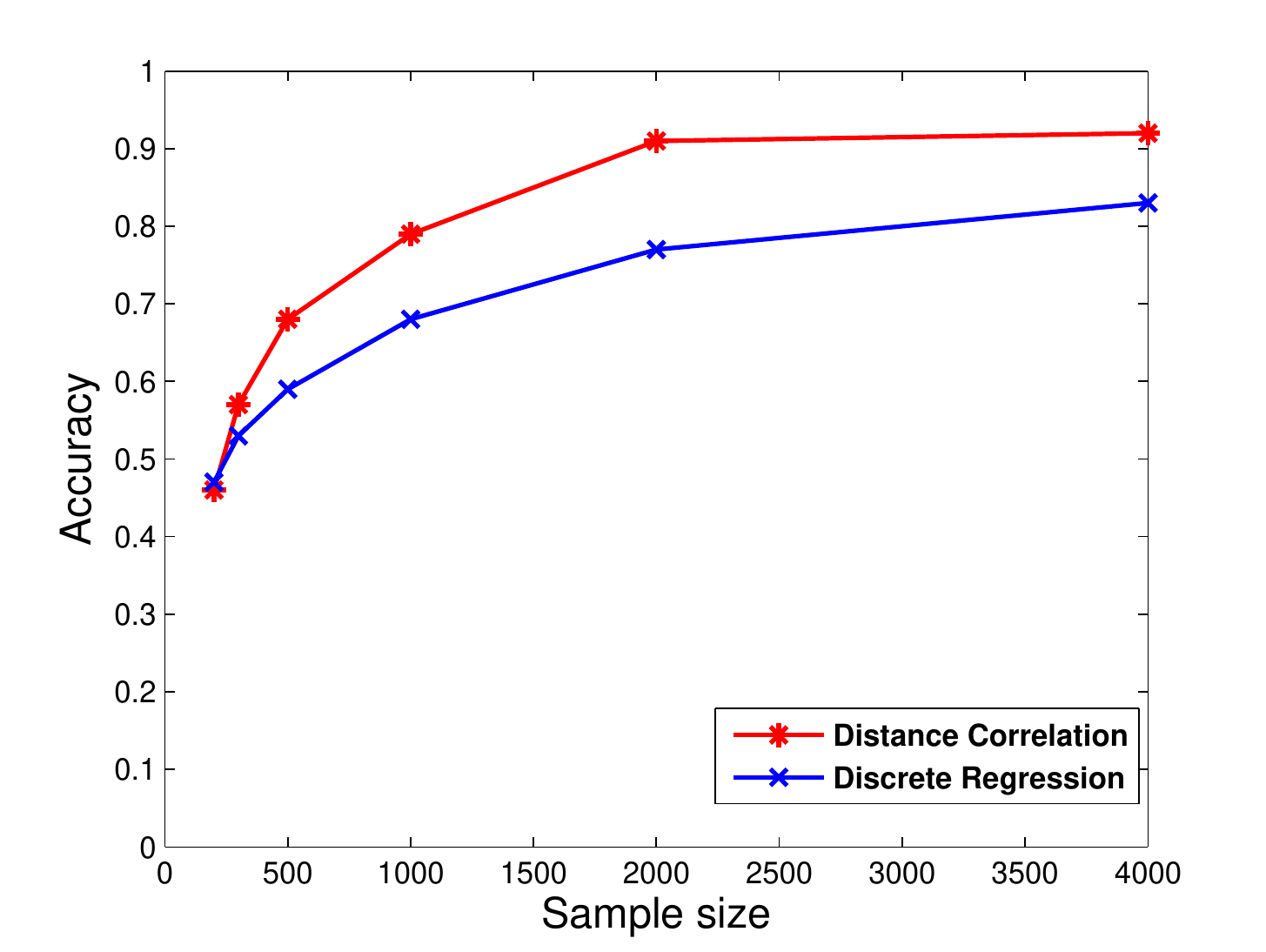} }}  
             \caption{Accuracy of the algorithms vs. sample size on different $\mathcal{N}$}
             \end{figure}
             
           From figure 2, one can see that DR performs slightly better than DC when $|\mathcal{N}|=2$. For example, when sample size is 4000, the accuracy of DR is 0.82 while that of DC is 0.78. We observe that ANMs with small $|\mathcal{N}|$ could sometimes yield small distance correlations in both directions. In these situations, the decision made by DC is close to a random guess. DC performs better than DR when $|\mathcal{N}|$ is larger.  The accuracies of DC become around 0.9 when sample size is large. But DC does not correctly identify all models. This is because the difference between the estimated distance correlations is sometimes small due to estimation random errors, and this may make the decision wrong.
           
                        \subsection{Models with Randomly Generated $P(X)$ and $P(Y|X)$} 
                         We now test the algorithms on the models with $P(X)$ and $P(Y|X)$ being randomly generated. To be specific, we generate $P(X)$  using the method in section 5.1. Then we generate $|\mathcal{N}|/4$ distributions on $\mathcal{Y}$ as a reference set. For  $x\in \mathcal{X}$, we generate $P(Y|x)$ by randomly taking one of the distributions in the reference set. We choose the domain size to be 
                          \begin{enumerate}
                           \item $(|\mathcal{X}|,|\mathcal{Y}|) = (12,12)$.
                           \item $(|\mathcal{X}|,|\mathcal{Y}|) = (15,15)$.
                           \item $(|\mathcal{X}|,|\mathcal{Y}|) = (18,18)$.
                          \item $(|\mathcal{X}|,|\mathcal{Y}|) = (20,20)$.
                          \end{enumerate}
                         For each setting, we generate 500 models. For each model, we sample 200, 300, 500, 1000, 2000, 4000 points, and apply DC and DR to them. The performance is showed in figure 3.  
                         \begin{figure}[h]%
                   \centering
                    \subfloat[ $(|\mathcal{X}|,|\mathcal{Y}|) = (12,12)$]{{\includegraphics[width=.5\textwidth]{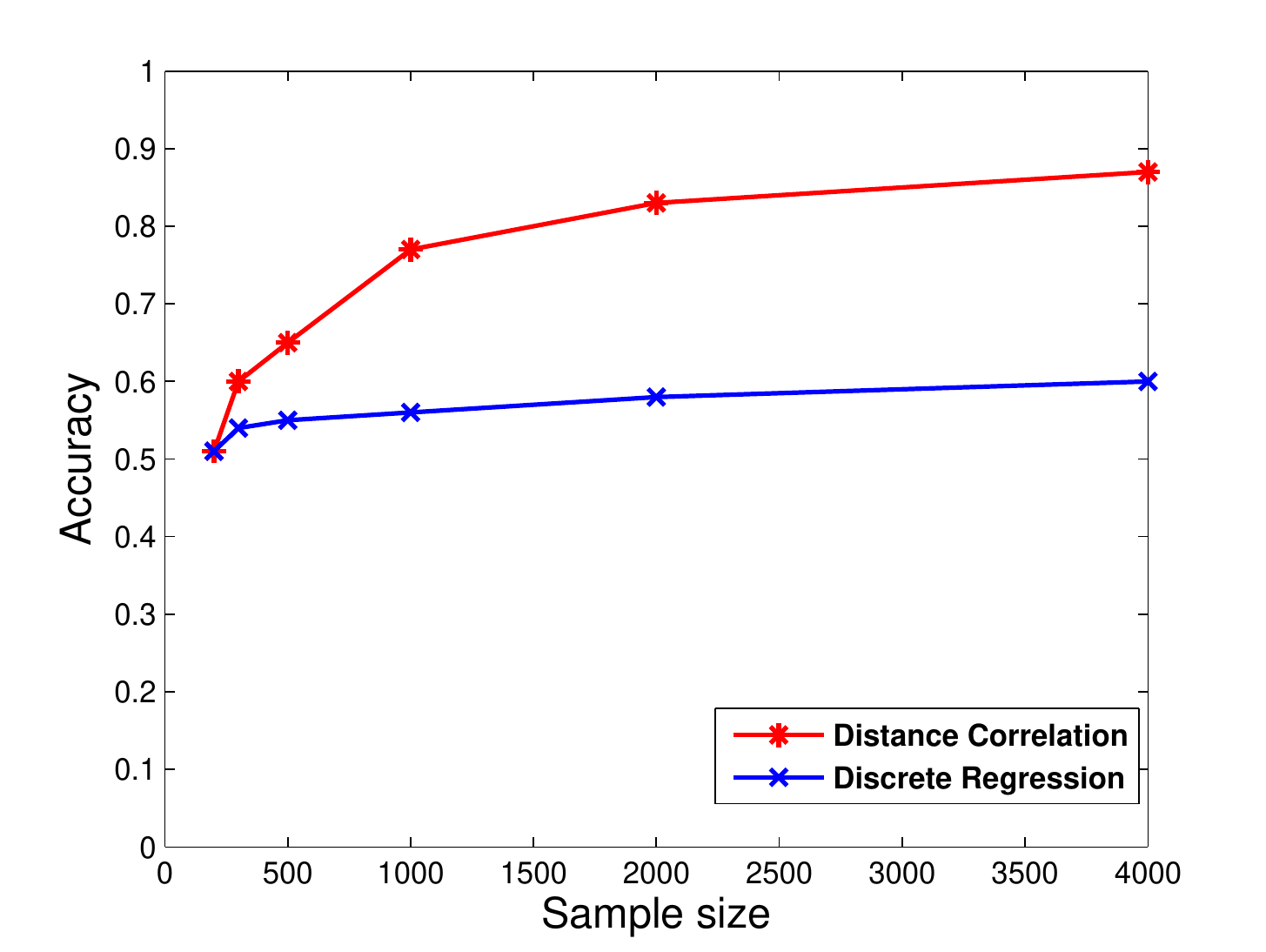} }}
                 \subfloat[ $(|\mathcal{X}|,|\mathcal{Y}|) = (15,15)$]{{\includegraphics[width=.5\textwidth]{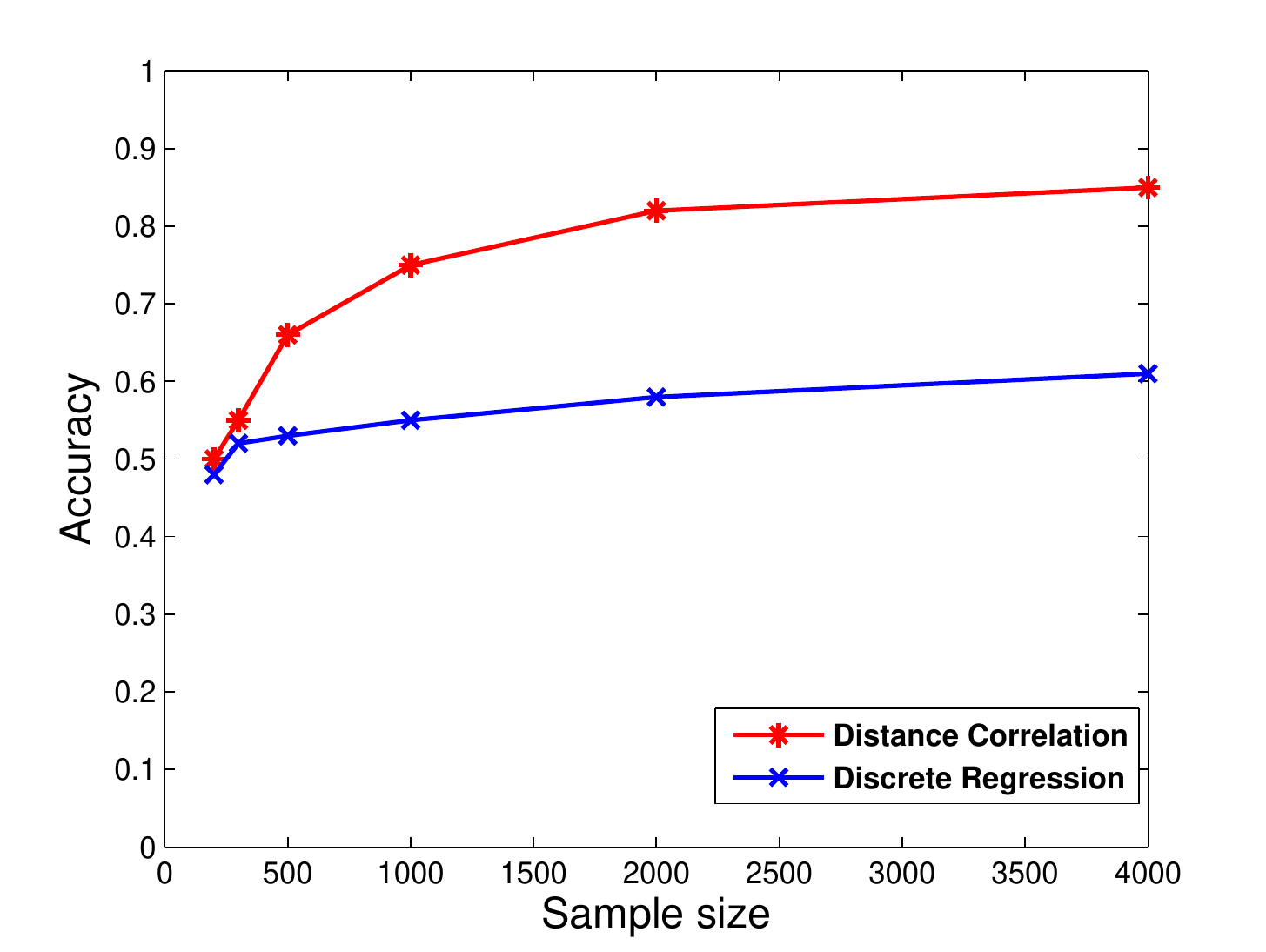} }}
                 
                 \subfloat[ $(|\mathcal{X}|,|\mathcal{Y}|) = (18,18)$]{{\includegraphics[width=.5\textwidth]{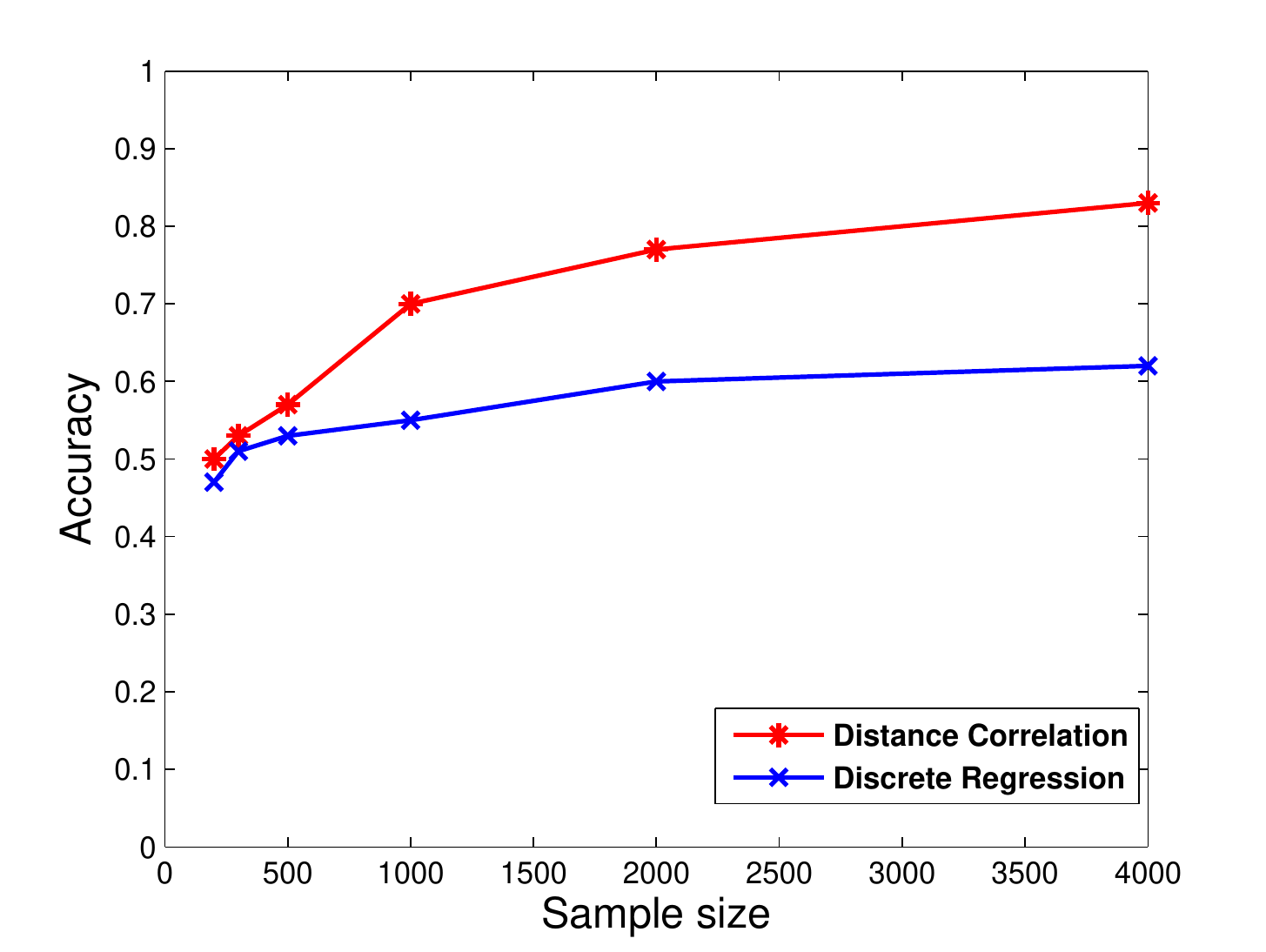} }}
                    \subfloat[ $(|\mathcal{X}|,|\mathcal{Y}|) = (20,20)$]{{\includegraphics[width=.5\textwidth]{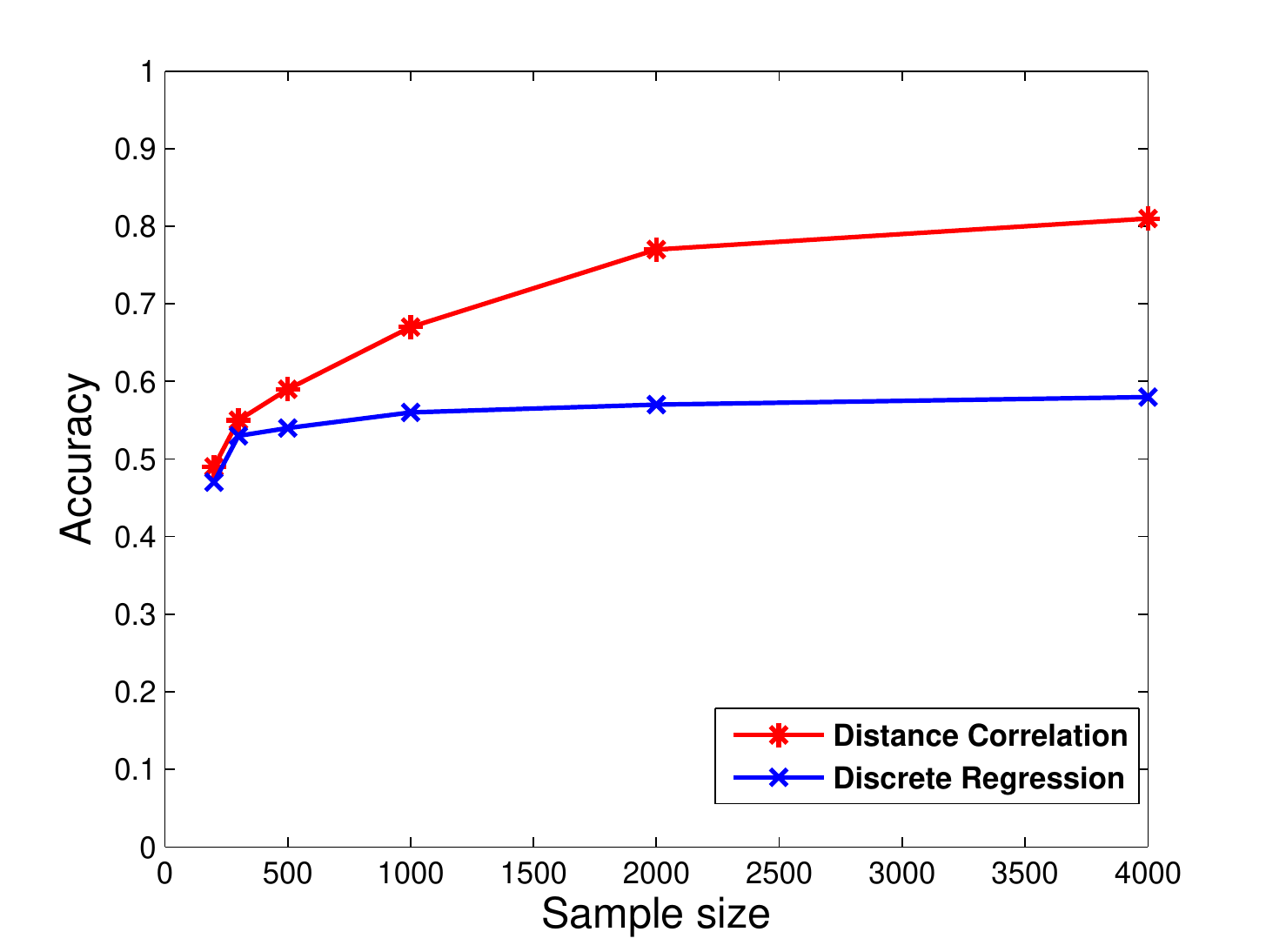} }}
                   \caption{Accuracy of the algorithms vs. sample size on different $(|\mathcal{X}|,|\mathcal{Y}|)$}
                    \end{figure}   
                      
             One can see that DR has unsatisfactory performance in these scenarios. This is because DR often makes a random guess since the models do not satisfy the ANM in either direction.     DC has satisfactory performance when sample size goes large. This shows that DC works in these scenarios while DR does not.  
             \subsection{On Efficiency}
             This section investigates the efficiency of the algorithms. We use the same experimental setting as that in section 5.2 (setting 2 and 4). For each sample size, we run them 100 times and record the total running time (seconds) of the algorithms. The records are showed in figure 4. 
              \begin{figure}[h]%
                                \centering
                              \subfloat[ $(|\mathcal{X}|,|\mathcal{Y}|) = (15,15)$]{{\includegraphics[width=.5\textwidth]{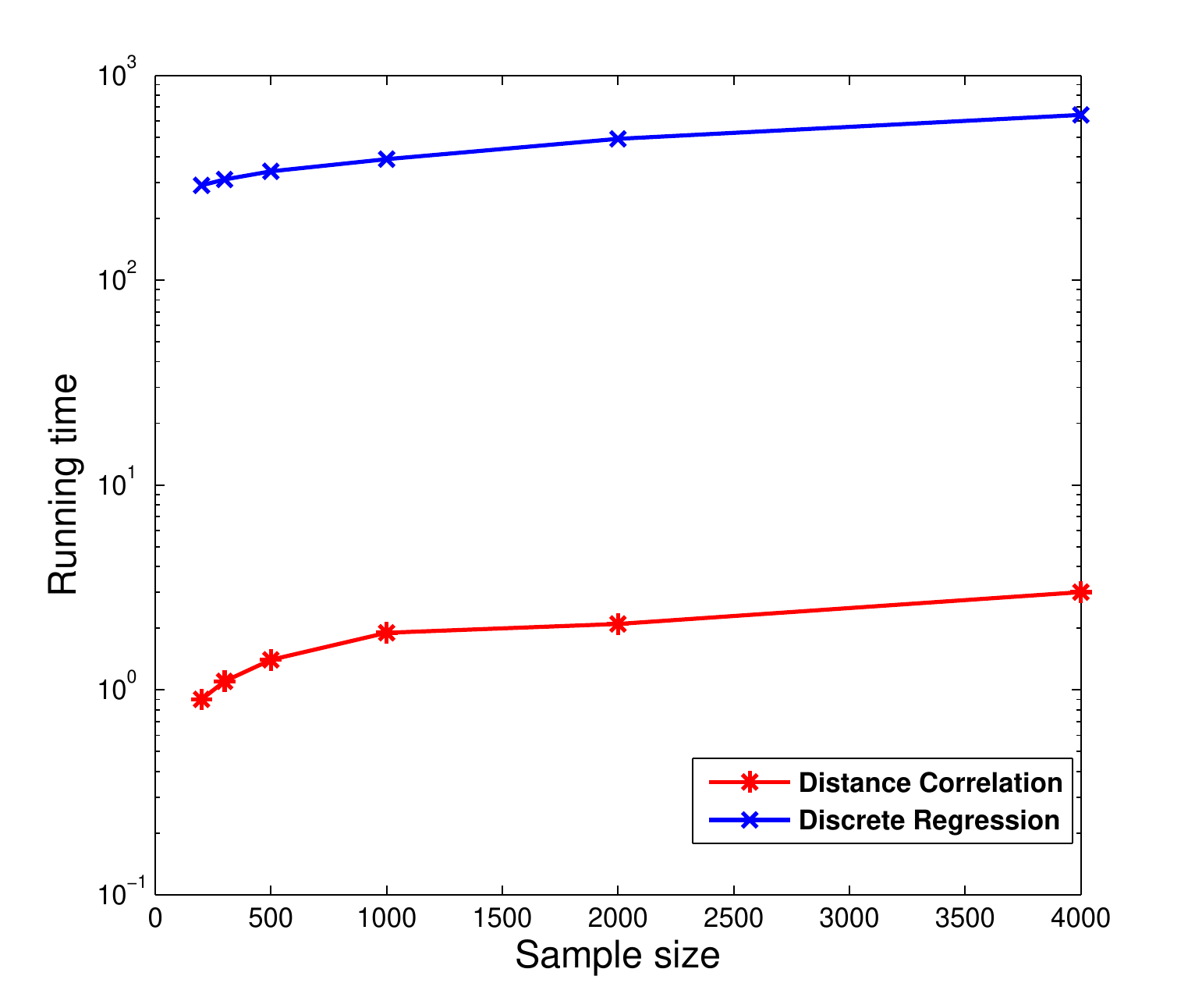} }}
                                 \subfloat[ $(|\mathcal{X}|,|\mathcal{Y}|) = (20,20)$]{{\includegraphics[width=.5\textwidth]{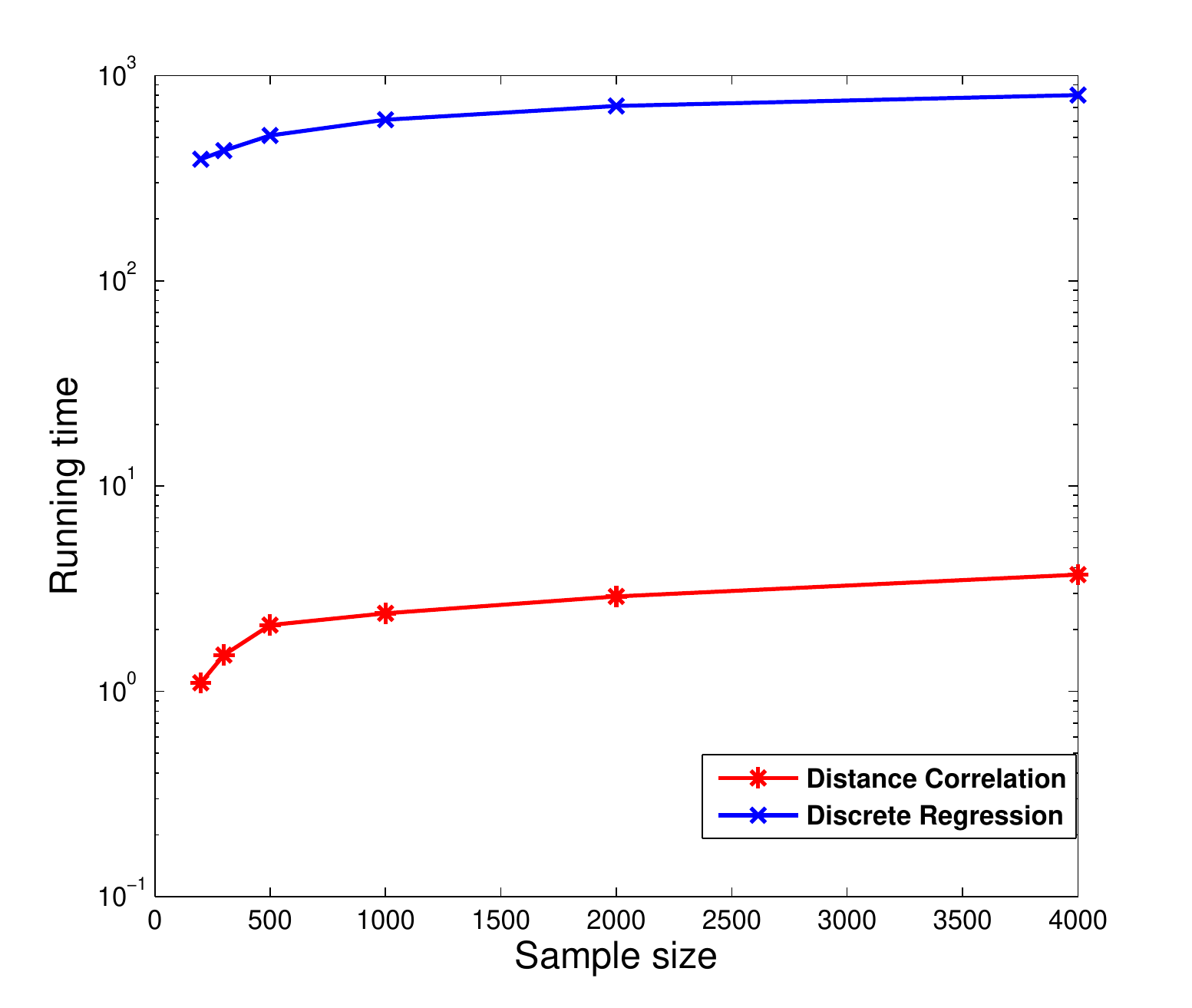} }}
                                \caption{Running time of the algorithms vs. sample size on different $(|\mathcal{X}|,|\mathcal{Y}|)$}
                                 \end{figure}    
                                 
  Figure 4 tells that DC has a higher efficiency than DR.  For example, when $(|\mathcal{X}|,|\mathcal{Y}|) = (15,15)$ and sample size is 1000, DR uses around 400 seconds to finish the experiments while DC uses only 2 seconds. This is because DR searches the whole domain iteratively to find a function that yields minimum dependence between residuals and regressors, which could be time-consuming in practice.   
    \subsection{On Parameter $\epsilon$}
               In the sections above, DC is forced to make a decision at each trial ($\epsilon = 0$). In this section, we examine the influence of $\epsilon$ on the performance of DC. This may help to set the values of $\epsilon$ in practice. We use the same experimental setting as that in section 5.2. The domain size is choose as $(|\mathcal{X}|,|\mathcal{Y}|) = (15,15)$ and $(|\mathcal{X}|,|\mathcal{Y}|) = (20,20)$. The sample size is fixed to be 4000.  We choose the parameter $\epsilon$ to be:
               \begin{enumerate}
            \item $\epsilon = 0.01$.
            \item $\epsilon = 0.05$.
           \item $\epsilon = 0.1$.
              \end{enumerate}
              For each setting, we generate 500 models and apply DC to them. The proportion of correctly identified models, proportion of wrongly identified models and proportion of non-identified models are showed in figure 5.
                \begin{figure}[h]%
                                  \centering
                                \subfloat[ $(|\mathcal{X}|,|\mathcal{Y}|) = (15,15)$]{{\includegraphics[width=.5\textwidth]{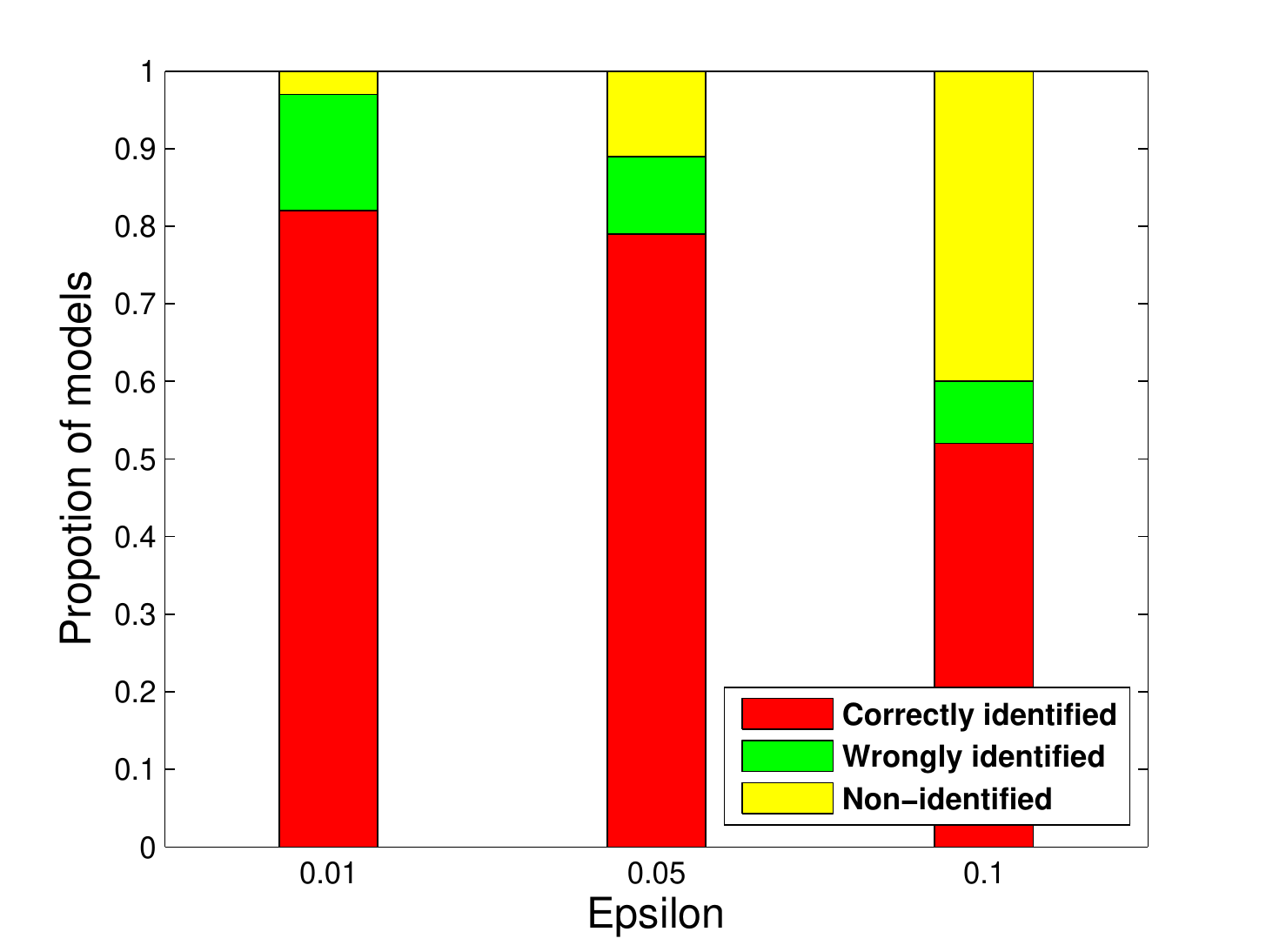} }}
                                   \subfloat[ $(|\mathcal{X}|,|\mathcal{Y}|) = (20,20)$]{{\includegraphics[width=.5\textwidth]{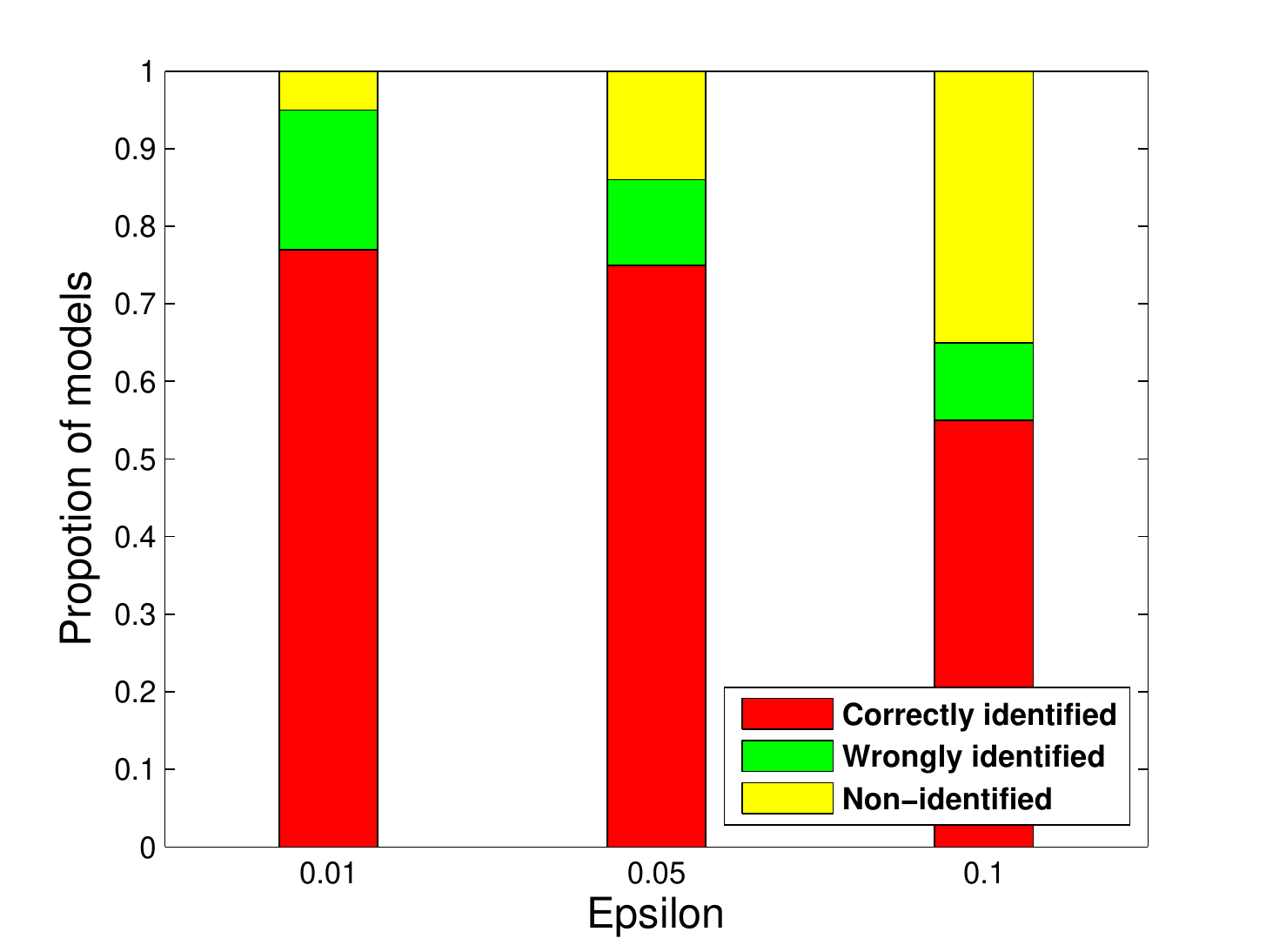} }}
                                  \caption{Proportion of models vs. threshold $\epsilon$ on different $(|\mathcal{X}|,|\mathcal{Y}|)$}
                                   \end{figure}    
                                   
   The proportion of non-identified models becomes large when the threshold is 0.1. This is because DC becomes conservative in this situation. The proportion of correctly identified models and that of wrongly identified models decrease as the threshold goes larger. However, the accuracy (the number of correctly identified models divided by the number of correctly identified models plus the number of wrongly identified models) increases. This is reasonable since the decisions made by DC under a higher threshold are more reliable. Based on the plotted results, we observe that the accuracy and decision date of DC are acceptable when $\epsilon=0.05$. We suggest that 0.05 is a reasonable choice for the parameter.  
   \subsection{On Decision Rates}
   In our algorithm, the threshold parameter $\epsilon$ controls the decision rates of DC. In other words, if we increase the $\epsilon$ (from 0 to 1), the decision rates decrease (from 100\% to 0\%).  In this section, we study the influence of the  parameter $\epsilon$ on the performance of DC. We follow the experimental setting in section 5.2, and we choose $(|\mathcal{X}|,|\mathcal{Y}|) = (15,15)$. We fix the sample size to be 5000, and we vary the decision rates (by changing the $\epsilon$ from 0 to 1). The plots showing the percentage of correct decisions versus the decision rate are available in figure 6. 
   \begin{figure}[h]
   \centering
   \includegraphics[width=0.5\textwidth]{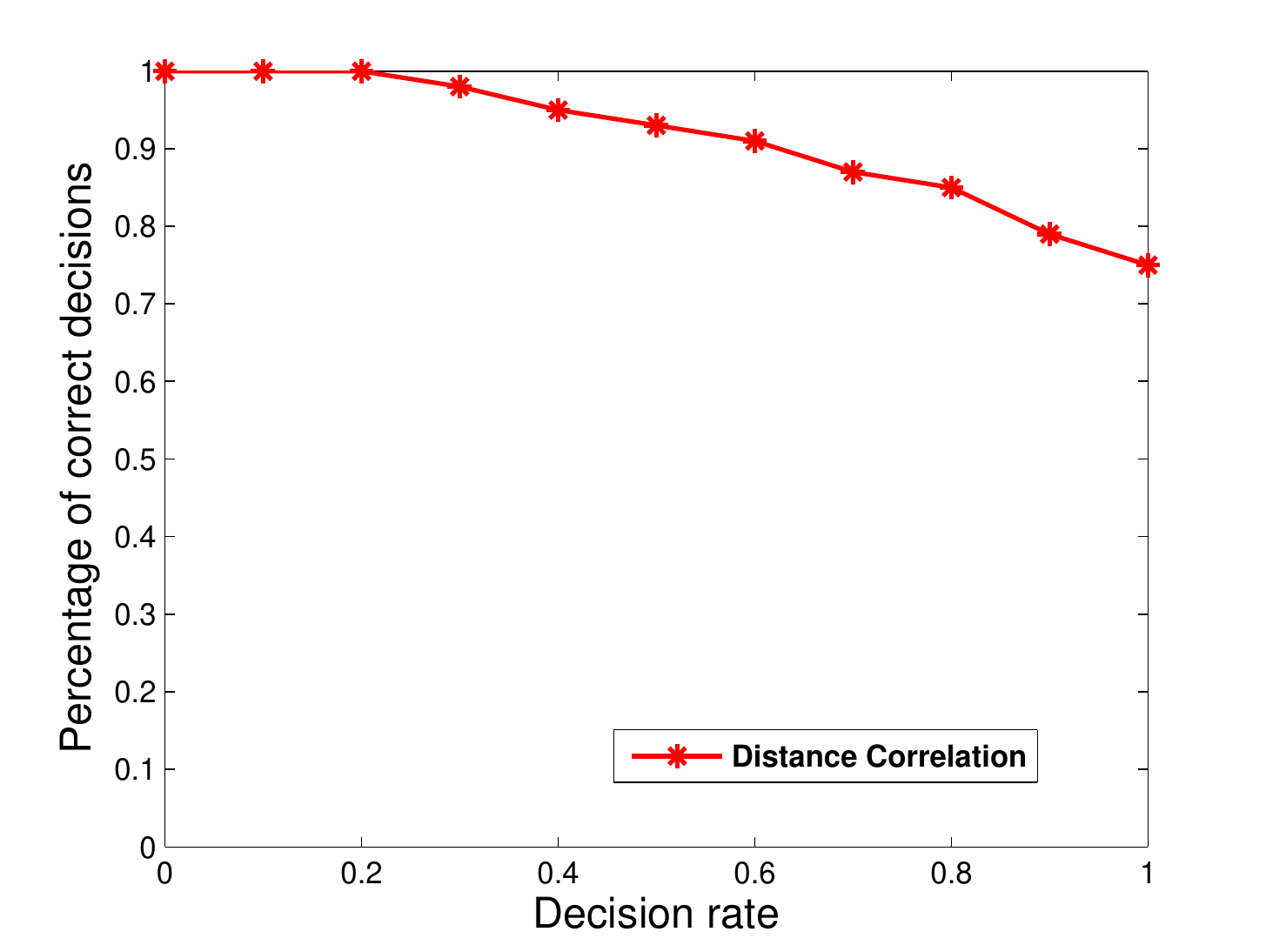}
   \caption{Percentage of correct decisions vs. decision rate}
   \end{figure}
   
   Obviously, the percentage of correct decisions decreases as the decision rate increases. For example, the percentage of correct decisions is 100\% when the decision rate is less than 20\%. But it becomes 77\% when the  decision rate is 100\%. This is acceptable since the decision would be more reliable if the algorithm makes decisions based on higher $\epsilon$. 
   \subsection{On Real World Data}
   We apply DC to real world benchmark cause-effect pairs\footnote{Available at https://webdav.tuebingen.mpg.de/cause-effect/}. The dataset contains records of 88 cause-effect pairs. We exclude the pairs numbered 17, 44, 45, 52, 53, 54, 55, 68, 71, 75.  They are either multivariate causal pairs or pairs that cannot be fitted into memory using our algorithm. We apply DC, DR\citep{peters2011causal}, IGCI\citep{janzing2012information}, LiNGAM\citep{lingam}, ANM\citep{hoyer2009nonlinear}, PNL\citep{zhang2009identifiability}, CURE\citep{sgouritsa2015inference}. The last 5 methods can be directly applied to the data. DC and DR could be applied to discretized data.   To make the variables discrete, we process them using the following method. For a variable $X$, if the maximum absolute value of all observations is less than 1, we process it by $round(20*X)$; else we process it by $round(X)$\footnote{For pair 65, 66, 67 which contain stock returns, we process the variables by $round(100*X)$}. For each pair, we generate 50 replicates using resampling techniques, and then apply the algorithms to the causal pairs. The boxplots showing the accuracies are in figure 7.    
   \begin{figure}[h]
      \centering
      \includegraphics[width=0.8\textwidth]{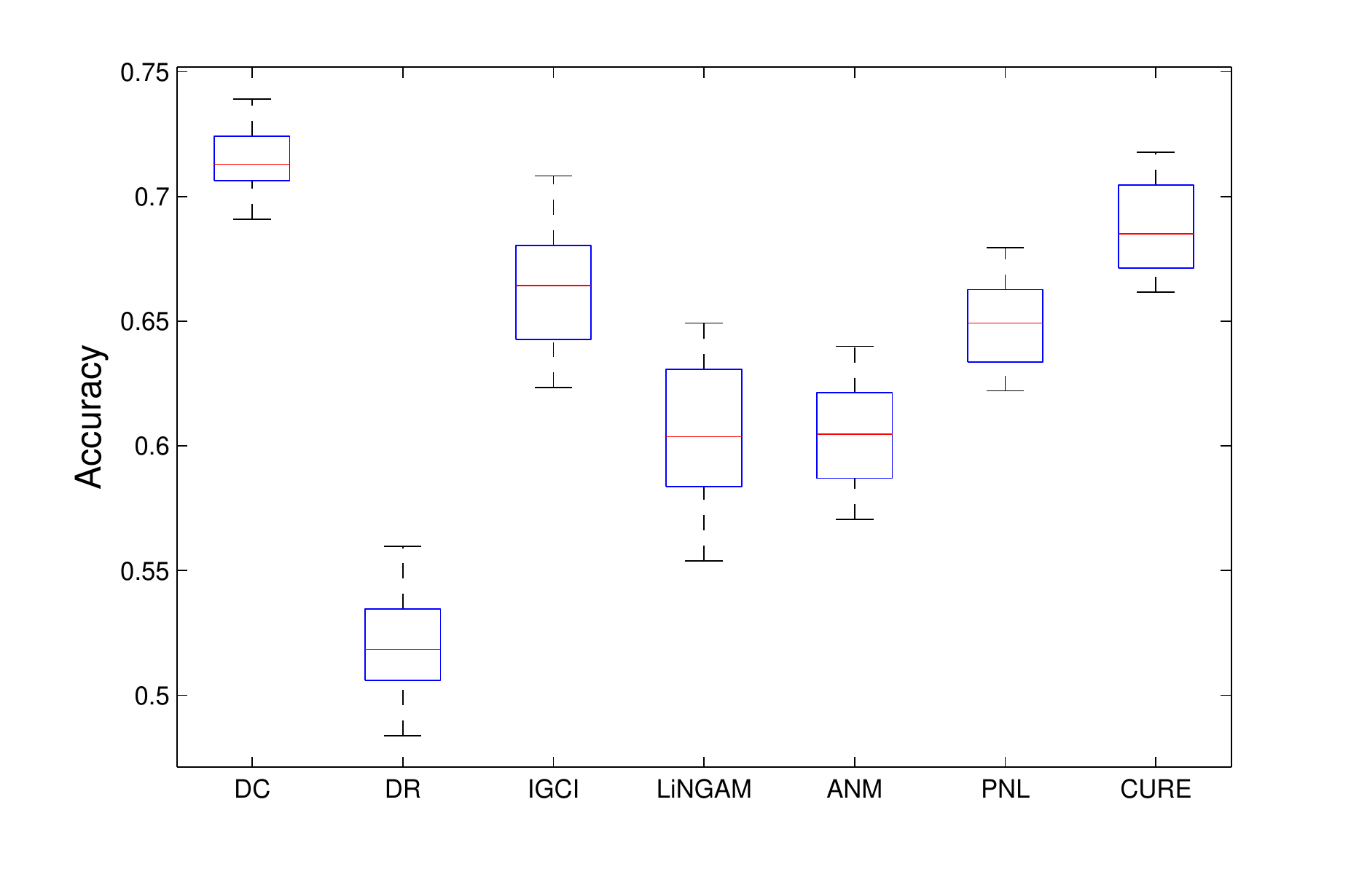}
      \caption{Accuracy of algorithms on 78 real world causal pairs}
      \end{figure}    
      
      Determining causal directions on real world data is challenging since the causal mechanisms are often complex and the data records could be noisy \citep{causality,spirtes2000causation}. However, figure 7 shows that DC has satisfactory performance on this task.  The average accuracy of DC is around 72\%, which is highest among all algorithms. ANM based methods (DR, LiNGAM, ANM) do not have a good performance. This may be because the assumptions of additive noise models restrict their applicability.    
      
      \section{Conclusions}
      In this paper, we deal with the causal inference problem on discrete data. We consider the distribution of the cause and the conditional distribution mapping cause to effect as independent random variables, and propose to discover the causal direction via estimating and comparing the distance correlations. Encouraging experimental results are reported. This shows that inferring the causal direction using the independence postulate is a promising research direction. In future we will try to  extend this method to deal with high dimensional  data. 
      
      \section*{Acknowledgments}
      The authors want to thank the editor and the anonymous reviewers for helpful comments. The work described in this paper was partially supported by a grant from the Research Grants Council of the Hong Kong Special Administration Region, China.
\bibliographystyle{apalike}
\bibliography{nc_paper}

%
%
%
%
\end{document}